\documentclass[journal]{IEEEtran}
\ifCLASSINFOpdf
\else
\fi
\usepackage{graphicx}
\usepackage{subfigure}

\usepackage{bbding}
\usepackage{amssymb}
\usepackage{multirow}
\usepackage{makecell}   
\usepackage{array}   
\usepackage{setspace}
\usepackage{booktabs}
\usepackage{amsmath} 

\usepackage{color}

\begin{document}
%
\title{Diversifying Topic-Coherent Response Generation for Natural Multi-turn Conversations}
%
%
%


\author{Fei~Hu,
        Wei~Liu,~\IEEEmembership{Member,~IEEE,}
        Ajmal~Saeed~Mian,~\IEEEmembership{Member,~IEEE,}
        Li~Li,~\IEEEmembership{Member,~IEEE}
\thanks{Manuscript received ***; revised ***. (\emph{Corresponding author}: \emph{Wei~Liu}.)}
\thanks{F. Hu is with the School of AI, Southwest University, Chongqing, China, and also with Network Centre, Chongqing University of Education, Chongqing, China, and with the School of Physics, Maths and Computing, The University of Western Australia, Western Australia, Australia (e-mail:
etz1@163.com).}
\thanks{W. Liu and A. S. Mian are with the School of Physics, Maths and Computing, The University of Western Australia, Western Australia, Australia (e-mail: wei.liu@uwa.edu.au; ajmal.mian@uwa.edu.au).}
\thanks{L. Li is with the School of Computer and Information Science, Southwest University, Chongqing, China (e-mail: lily@swu.edu.cn).}
}

%
%

\markboth{Journal of \LaTeX\ Class Files,~Vol.~14, No.~8, August~2019}%
{Shell \MakeLowercase{\textit{et al.}}: Bare Demo of IEEEtran.cls for IEEE Journals}
%



\maketitle

\begin{abstract}
Although response generation (RG) diversification for single-turn dialogs has been well developed, it is less investigated for natural multi-turn conversations. Besides, past work focused on diversifying responses without considering topic coherence to the context, producing uninformative replies. In this paper, we propose the Topic-coherent Hierarchical Recurrent Encoder-Decoder model (THRED) to diversify the generated responses without deviating the contextual topics for multi-turn conversations. In overall, we build a sequence-to-sequence net (Seq2Seq) to model multi-turn conversations. And then we resort to the latent Variable Hierarchical Recurrent Encoder-Decoder model (VHRED) to learn global contextual distribution of dialogs. Besides, we construct a dense topic matrix which implies word-level correlations of the conversation corpora. The topic matrix is used to learn local topic distribution of the contextual utterances. By incorporating both the global contextual distribution and the local topic distribution, THRED produces both diversified and topic-coherent replies. In addition, we propose an explicit metric (\emph{TopicDiv}) to measure the topic divergence between the post and generated response, and we also propose an overall metric combining the diversification metric (\emph{Distinct}) and \emph{TopicDiv}. We evaluate our model comparing with three baselines (Seq2Seq, HRED and VHRED) on two real-world corpora, respectively, and demonstrate its outstanding performance in both diversification and topic coherence.
\end{abstract}

\begin{IEEEkeywords}
Response Generation Diversification, Topic Coherence, Natural Multi-turn Conversation, Multi-mode Distribution, Recurrent Neural Nets.
\end{IEEEkeywords}

%
\IEEEpeerreviewmaketitle

\section{Introduction}
\IEEEPARstart{R}{esponse} generation (RG) has been playing an increasing important role in Natural Language Generation (NLG) as it draws close to industry manufacture and our daily life. Neural net models building upon encoder-decoder learning \cite{sutskever2014sequence,cho2014learning} have been demonstrated effective in RG and have achieved a lot of success \cite{rush2015neural,mou2016sequence,yan2016shall,tian2017make}, while these models suffered from \emph{safe reply} problem \cite{li2015diversity,wu2018neural} as they prefer producing generic and safe replies like ``thank you'' and ``I am sorry'', and high-frequent function words like ``the'' and ``no'' due to the high frequency of these patterns and words in the training data. Although these generic responses are helpful to promote the results in terms of accuracy, they are less informative and even meaningless to the post. In addition, accurate replies are not good answers because we would like to respond based on contextual semantics and conversational environments rather than based on an accurate-reply handbook. Diversifying the responses will make conversations more informative, more interesting, and more like human interaction.

\emph{Safe reply} problem is a big challenge in RG. While encoder-decoder models follow the functional principle of \cite{ritter2011data}, making both the source sentence and target sentence subject to the same latent variables like the Machine Translation (MT) does, this principle neglects the intrinsic difference between MT and RG that MT treats the sentence pairs of the same meanings but RG has to move further to a richer response rather than the post \cite{pei2018s2spmn}. Besides, encoder-decoder models deal with RG based on post-response pairs of sentences that narrows the distribution of predicted responses and gives those highly frequent words and patterns a higher chance to show themselves in the final generation \cite{pei2018s2spmn,kassarnig2016political}. Many works have proposed to mitigate the \emph{safe reply} problem, producing informative and interesting replies \cite{li2015diversity,perez2005application,deriu2017end,deriu2018syntactic,elder2018e2e,wen2015semantically,novikova2017e2e,hu2017toward,sammut2010beam,vijayakumar2016diverse,asghar2016deep,cibils2018diverse,vijayakumar2018diverse,du2018variational,xing2017topic,baheti2018generating,wang2018learning,le2018variational} . These works are helpful to some extent, as \cite{deriu2017end,deriu2018syntactic,elder2018e2e} learning from several target references for each post to broaden the generation distribution, \cite{sammut2010beam,vijayakumar2016diverse,asghar2016deep,cibils2018diverse,vijayakumar2018diverse} producing a set of diversified candidate replies, and \cite{xing2017topic,baheti2018generating,wang2018learning} leveraging the topic distribution to bias final responses. However, these methods only work on single-turn dialog tasks.

Recent years, Sequence-to-Sequence model (Seq2Seq) \cite{luong2015effective} has demonstrated its effect in modeling multi-turn dialogs \cite{vinyals2015neural}. It is based on encoder-decoder framework which encodes the sequence of tokens recurrently. Seban et. al extended Seq2Seq to model relationship between utterances, proposing the Hierarchical Recurrent Encoder-Decoder model (HRED) which made the final responses more comprehensive and informative \cite{serban2016building}. After that, a lot of works were proposed to model correlations between utterances, producing diverse responses \cite{zhao2017learning,serban2016generative,shen2017conditional,serban2017hierarchical}. Unfortunately, these diverse responses are not topic-related to the context due to the lack of topic information.

In this paper, we build a Seq2Seq framework \cite{luong2015effective} and extend the functional principles with respect to both variational methods \cite{serban2017hierarchical,du2018variational} and topic methods \cite{xing2017topic,baheti2018generating,sohn2015learning}, proposing both global and local strategies that inject the global contextual information and the local topic information into the response for multi-turn conversations. The key idea of this paper is to model both the global contextual distribution and the local topic distribution, and to train them jointly. It is like the way of real-world conversations that people generalize the context based on the previous turns of talks and replace the responses¡¯ patterns with semantically (topic-) similar ones to make the conversation more informative and interesting.

Global contextual distribution implies linguistic rules. We resort to the latent Variable Hierarchical Recurrent Encoder-Decoder model (VHRED) \cite{serban2017hierarchical} to learn it where Conditional Variational Auto-Encoder (CVAE) \cite{sohn2015learning,pandey2016variational} is used to acquire the knowledge of speaking skills, gaining correlations between utterances. We leverage the discourse-level knowledge to help produce more comprehensive responses.

Local topic information is explicitly sampled from a topic distribution where words have correlation probabilities over a series of topics. Specifically, we firstly build a sparse matrix which distributes topics over all non-functional words in the vocabulary, i.e., each topic is denoted by a word. Then we extend the word-level topic distribution to generalize higher-level topics into a dense topic matrix using Non-negative Matrix Factorization (NMF) \cite{lee1999learning}, i.e., each topic turns into a high-level pattern. Thus, topic values sampled from this dense topic matrix could enrich word-level expression in the final generation.

Global contextual information and local topic information are both dynamic because they are conditioned on dynamic context within various dialogs. As a result, patterns in responses are diversified without deviating the informativeness to the context.

We study RG for open-domain and multi-turn conversation systems because they are in accordance with real-world scenes and are more challenging than task-oriented \cite{Reddy2018Multi} and single-turn \cite{pei2018s2spmn} conversation systems. In daily lives, people talk to each other in more than one utterance, and previous utterances contain contextual information that could be used to support and remain following conversations. In open-domain and multi-turn RG, \emph{safe reply} is much more an issue because the long and redundant contextual utterances bring more functional patterns. Therefore, traditional encoder-decoder models cannot learn multi-turn utterances effectively rather than single-turn and short conversations.

In summary, our contributions are as follows:
\begin{enumerate}
\setlength{\abovecaptionskip}{0.cm}
\setlength{\belowcaptionskip}{-0.cm}
  \item We use bias factors from two separated distributions (global contextual distribution and local topic distribution) to influence dull responses, producing diversified yet topic-related replies.
  \item We diversify RG in the dialog-level and word-level, respectively.
  \item We advocate an explicit metric (\emph{TopicDiv}) to measure the topic divergence between the post and the according response. In addition, we combine the diversification metric (\emph{Distinct}) and \emph{TopicDiv} to propose an overall metric (\emph{F} score) which does the comprehensive evaluation of diversification and topic coherence.
\end{enumerate}

In this paper, we introduce two multi-turn dialog datasets, Daily Dialogs \cite{li2017dailydialog} and Ubuntu Dialogs \cite{lowe2015ubuntu}, to evaluate our model. Daily Dialogs is less noisy, in which the dialogues are well organized and carefully selected from human-written communications, reflecting our daily communication way and covering various topics about our daily life. Ubuntu Dialogs is two orders of magnitude bigger than Daily Dialogs, containing almost one million two-person conversations which were extracted from the Ubuntu chat logs, being used to receive technical support for various Ubuntu-related problems. We also compare to three state-of-the-art models: SEQ2SEQ \cite{vinyals2015neural}, HRED \cite{serban2016building} and VHRED \cite{serban2017hierarchical}. Experimental results show that our model significantly outperforms the other three models in generating diversified and topic-coherent responses.

\section{Related work}
Diversifying RG has been attracting a growing number of researchers, unyielding to the demands to match the target reference replies, and turning up unusual results. Traditional RG diversification methods are roughly divided into two categories: task-oriented (or data-driven) methods and open-domain methods. While task-oriented methods only work with elaborate corpora \cite{deriu2017end,deriu2018syntactic,elder2018e2e} and extra carefully selected supplementary data \cite{Reddy2018Multi}, open-domain methods are flexible in real-world environments, such as mutual information methods \cite{li2015diversity}, beam search methods \cite{sammut2010beam,vijayakumar2016diverse,asghar2016deep,cibils2018diverse,vijayakumar2018diverse}, topic bias methods \cite{baheti2018generating,wang2018learning} and variational methods \cite{du2018variational,le2018variational}. These methods only work with single-turn dialog tasks while multi-turn conversations were not well studied till the model of Seq2Seq \cite{luong2015effective}.

Seq2Seq is a recurrent encoder-decoder model. It leverages recurrent nets to encode the context into a fixed-size vector which is then used to decode the output response. Vinyals and Le have broken the logjam for modeling multi-turn conversations by using the Seq2Seq model \cite{vinyals2015neural}. They utilized Long Short-Term Memory (LSTM) as the Encoder and the Decoder, respectively, encoding previous multiple utterances in a compressed vector and decoding it to produce the output response. However, Seq2Seq cannot learn lengthy dialogs effectively due to the natural flaws of vanishing memory with recurrent models (including LSTM) when encoding long past information \cite{pascanu2013difficulty,hochreiter1998vanishing}. Moreover, the problem of vanishing long-term memory confines the model to a short range of the later tokens, dampening learning language's multi-mode distributions which might exist in the far-previous contextual segments.

In order to learn language patterns effectively and comprehensively, Seban et al extended Seq2Seq to propose HRED \cite{serban2016building} by incorporating an additional recurrent net to model correlations between utterances. In this way, long-term language patterns are encoded in a compressed vector. This compressed vector certainly implies dialog-level contextual information and turns out to diversify the generated responses. Specifically, HRED generalized contextual information and made great use of it to bias the safe replies. The generalized contextual information makes up the deficiency of the lack of long-term contextual information.

Considering the successful variational methods in modeling natural language \cite{chung2015recurrent}, CVAE was used to improve modeling multi-turn conversations \cite{serban2017hierarchical,zhao2017learning,serban2016generative,shen2017conditional} and has demonstrated its effects in diversifying generated responses \cite{zhao2017learning,serban2016generative,shen2017conditional}. Seban et al extended HRED to propose VHRED \cite{serban2017hierarchical} by incorporating a latent distribution (instead of the compressed vector in HRED) to model correlations between utterances. The latent distribution is learned by using CVAE, which leveraged diverse contexts as conditional factors to dynamically model the correlational knowledge between utterances.

Traditional diverse RG systems for natural multi-turn conversations have improved the encoder-decoder model by deviating the final response from the target reference replies, which, however, either do not satisfy the multi-distribution quality of the language given the syntactically and semantically diverse context, or lack the topic information related to the context.

CVAE aims to encode the knowledge between utterances into a high-level data distribution. And conditioned on the diverse context, its distribution becomes dynamic \cite{dykeman2016cvae}. We extend VHRED \cite{serban2017hierarchical} (which uses CVAE) to learn dynamic distribution in discourse level. Meantime, a pre-trained topic matrix provides word-level dynamic distribution given the conditional words in the context. Both the discourse-level (global) and word-level (local) information foster the system to produce interesting and informative responses.

\section{Methodology}
\subsection{Overview}
In multi-turn conversational systems, a dialogue can be considered as a sequence of utterances. And each utterance contains various length of tokens. Formally, we have $D=\{U_1,\cdots,U_M\}$ and $U_m=\{w_{m,1},\cdots,w_{m,N_m}\}$, where $D$ is a dialogue, $U_m$ is the $m$-th utterance of $D$, and $w_{m,n}$ is a token at position $n$ of $U_m$. The RG task is to predict $U_m$ given the previous contextual utterances $\{U_1,\cdots,U_{m-1}\}$. The prediction process is formulated as follows:
\begin{equation}\label{eq1}
\begin{split}
  p(U_m)&=p(w_{m,1},\cdots,w_{m,N_M}) \\
  &=\prod_{n=1}^{N_m}p(w_{m,n}\big|w_{m,<n};U_{<m})
\end{split}
\end{equation}

From this formulation we can see, the RG prediction counts on two parts: $U_{<m}$ and $w_{m,<n}$. That is, the RG system models the prediction with a two-level hierarchy: a sequence of utterances, and the tokens in current utterance \cite{serban2016building}.

Overall, our work is a Seq2Seq model \cite{luong2015effective}, which is known as the recurrent encoder-decoder model \cite{shang2015neural}. As a prevalent neural machine translation approach, Seq2seq has been successfully applied to RG \cite{sordoni2015neural,vinyals2015neural}. In particular, Seq2seq is used to learn the embeddings of the context of the previous utterances to generate tokens in the current utterance. Seq2Seq improves RG in terms of accuracy, producing standard replies adhering to the reference replies, but failing to address the \emph{safe reply} problem.

In order to mitigate the \emph{safe reply} problem, we leverage both the global contextual and the local topic offsets to bias the generic replies. Specifically, we resort to VHRED \cite{serban2017hierarchical} to learn the global contextual offset and leverage NMF \cite{lee1999learning} to learn the local topic offset, proposing the Topic-coherent Hierarchical Recurrent Encoder-Decoder model (THRED) to produce not only diversified but also topic-coherent replies.

The VHRED has demonstrated an ability to improve diversification of RG \cite{zhao2017learning,serban2016generative,lowe2017towards,shen2017conditional}. In this paper, we resort to the VHRED to learn the global contextual information of dialogs which utilizes the CVAE to learn contextual structures and correlations between utterances within each dialog, learning common linguistic rules. The global linguistic knowledge is injected into a global contextual distribution. And then, conditioned on the contextual utterances in the dialog, a latent variable \emph{z} is sampled from the distribution. It encodes the global linguistic knowledge which involves the context of current dialog, improving the decoder to produce a more comprehensive reply.

Besides, we use NMF to learn the local topic information conditioned on the words in current dialog. The global context information reflects general knowledge of the dialog, while the local topic information implies the topics of all words in the dialog. The two offsets do not simply change patterns in the generated response, but improve the response to reserve speaking skills and linguistic rules, and to follow the topics of the context.

\begin{figure}
  \centering
  \includegraphics[scale=0.38]{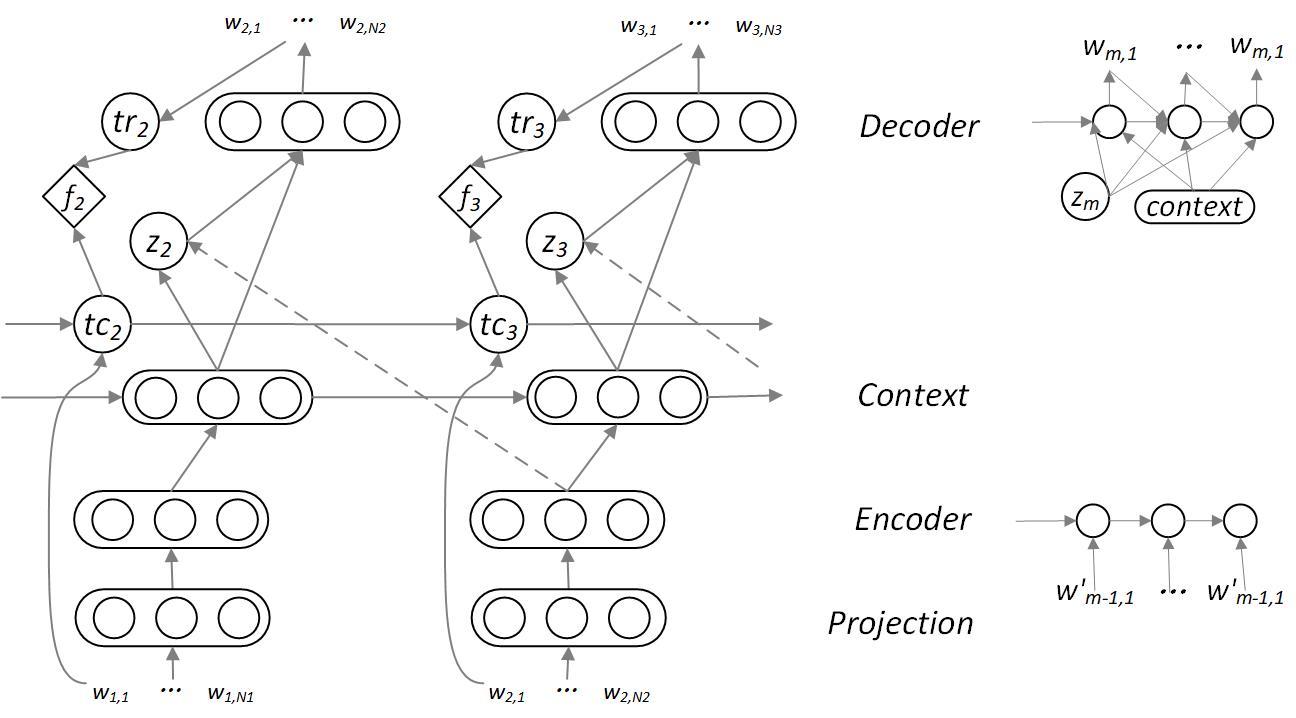}\\
  \caption{The structure of the Topic-coherent Hierarchical Recurrent Encoder-Decoder model (THRED).}\label{fig1}
\end{figure}

As shown in Fig. \ref{fig1}, in the left is the framework we proposed where $z$ is the latent variable of the global context distribution, $tr$ and $tc$ are sampled from the local topic distribution conditioned on the tokens of the context and the tokens of the predicted response, respectively. The subscripts represent time step of utterances in the dialog. The proposed framework has four layers: \emph{Projection}, \emph{Encoder} that is depicted in the right bottom, \emph{Context}, and \emph{Decoder} that is depicted in the right top. The \emph{Projection} is a full-connected neural net, encoding tokens into dense embeddings with the same dimensional size of the following layers. The \emph{Encoder} is a recurrent net which sequentially encodes the token embeddings of the utterance, learning utterance-level information. The \emph{Context} is also a recurrent net which encodes the temporal utterances of the dialog, learning dialog-level information. The \emph{Decoder} encodes both the context embedding and the latent variable z, producing temporal sequence of the response. Meantime, the function $f$ measures the distance between $tc$ and $tr$, which is an optimization constraint to bias the model to learning topic correlations between the replies and the context.

The \emph{Encoder} is a bidirectional LSTM net \cite{hochreiter1997long}. The \emph{Context} and the \emph{Decoder} are unidirectional LSTM nets. In the following subsections, we will explain the learning processes of the $z$ distribution (global contextual distribution) and the $t$ distribution (local topic distribution including $tr$ and $tc$), respectively.

\subsection{Learning global contextual distribution}
The global contextual distribution learns the discourse-level knowledge of conversations. To this end, we resort to VHRED \cite{serban2017hierarchical} which utilizes the CVAE \cite{sohn2015learning,pandey2016variational} to simulate a discourse-level distribution. And we build the discourse-level knowledge by sampling from this distribution when predicting the response.

CVAE improved the Variational Auto-Encoder (VAE) model \cite{kingma2013auto} by introducing a conditional factor. Vanilla VAE encodes all data into a single-mode distribution no matter the different patterns of these data, while CVAE encodes data with different conditional factors into respective distributions. The conditional factor $c$ is a prior knowledge, and another posterior factor $X$ is introduced which can be taken as the label of the according sample. Thus, optimizing CVAE can be thought of as a supervised learning that expects the target $X$ conditioned on $c$. After having learned the variational distribution, we can sample dynamic patterns from it conditioned on different conditional factors. We formulate the $c$-conditioned objective function as:
\begin{equation}\label{eq2}
\begin{split}
  Obj_{global}=&-E_{z\sim Q(z|X,c)}\big[\log P(X|z,c)\big] \\
  &+KL\big[Q(z|X,c)||P(z|c)\big]
\end{split}
\end{equation}
Where $z$ is the latent variable sampled from $Q$, $E_{z\sim Q(z|X,c)}$ expects all samples in the distribution $Q$, $P$ is the prior distribution which approximates the posterior distribution $Q$, $ KL\big[\big]$ is the Kullback-Leibler divergence function which is used to measure how one probability distribution is different from another one. Optimization is performed by minimizing the lower bound of this objective function, i.e., $-E_{z\sim Q(z|X,c)} \big[\log P(X|z,c)\big]$, while the $KL$ divergence is greater than zero in all time. $KL\big[Q(z|X,c)||P(z|c)\big]$ ensures $P$ approximates $Q$ since $X$ is not available in the inference step and $z$ can be sampled from $P$ instead of $Q$.

$z$ (in Eq. \ref{eq2}) is of the discourse-level distribution. The learning process of $z$ in the training step and the sampling process of $z$ in the testing step are detailed as follows: In the training step, $Q(z|X,c)$ encodes both the previous utterances $c$ and the following expected utterance $X$ to model $z$. Since $Q(z|X,c)$ considers the whole dialog (the previous utterances plus the expected replied utterance), it therefore learns the exactly accurate discourse-level knowledge. The expected utterance $X$, as a posterior factor, is not available in the testing step, thus, another distribution $P(z|c)$ is introduced to take the place of $Q(z|X,c)$. $P(z|c)$ models a prior distribution which only considers the previous utterances $c$. By using the $KL$ divergence function as a regularization term (see Eq. \ref{eq2}), $P(z|c)$ approximates $Q(z|X,c)$. In the testing step, the prior distribution $P(z|c)$ instead of $Q(z|X,c)$ is used to fill the gap between the expected response and the discourse-level knowledge in dialogs.

In this paper, we encode $z$ in a $d_z$-length vector. Both $Q$ and $P$ are Gaussian distributions. $Q=\mathcal N(\mu_{posterior},v_{posterior})$ and $P=\mathcal N(\mu_{prior},v_{prior})$ where mean $\mu$ and covariance $v$ are encoded in $d_z$-length vectors, respectively.

\subsection{Learning local topic distribution}
\subsubsection{Building the topic matrix}
The local topic distribution explicitly encodes topics of all non-functional words in a topic matrix. In particular, we utilize PPMI \cite{turney2010frequency} to build a sparse word-topic matrix where each topic is a word in the vocabulary. Then we use NMF \cite{lee1999learning} to factorize it to obtain a dense word-topic matrix where each topic turns into a high-level pattern.

By using PPMI, we construct a high-dimensional matrix $M\in \mathbb{R}^{|V_w|\times|V_w|}$ where the row denotes the list of words and, the column represents the list of contextual features. Both the row and the column are the list of non-functional words in the vocabulary. The value of the matrix cell $M_{ij}$ is the PPMI value that suggests the associated relationship between the word $w_i$ and the contextual feature $k_j$, which can be estimated by:
\begin{equation}\label{eq3}
  M_{ij}=\max\Big \{\log \Big (\frac{p(w_i,k_j)}{p(w_i )p(k_j)}\Big ),0\Big \}
\end{equation}
Where the $\max$ function ensures that only positive correlations of word-feature pairs are reserved and negative correlations are ignored by setting them zero.

The sparse PPMI matrix raises two issues: 1) The topic representation (i.e., word-level representation of topics) is too specific to be adaptive in learning stable topic distribution; 2) The sparsity results in both excessive memory consumption and extreme time complexity when training the model. In order to mitigate the sparsity problems, we resort to NMF to cluster sparse topics in dense topic patterns. NMF factorizes the sparse PPMI matrix $M$ into two dense matrices $W$ and $H$, mathematically abstracting it as $M\approx WH$. The approximation of $M$ is achieved by minimizing the objective function $\min\limits_{W,H} ||M-WH||_F$. $W$ is a $|V_w| \times p$ matrix, and $H$ is a $p \times |V_w|$ matrix then $p$ can be significantly less than $|V_w|$ (in this paper, we set $p=40$).
Unlike the Singular Value Decomposition (SVD) which might generate negative values in the final dense matrix, the $W$ produced by NMF has only positive elements, i.e., correlated topic patterns are reserved yet uncorrelated patterns are ignored. The non-negative quality guarantees that the dense word-topic distribution conforms to the sparse word-topic distribution, remaining the positive relationship between words and topic features.

From the training logs, we randomly selected ten topic divergence values at ten successive training epoches with PPMI and NMF, respectively, listing them in Table \ref{tb1} 
, where the topic divergence value is the $KL$ divergence between the context and according predicted response. The two ranges are of the same number of training epoches. The variance values for the bunch of topic divergence values with PPMI and the bunch of topic divergence values with NMF are calculated, respectively. As we can see, NMF has a much smaller variance value, i.e., the dense topic patterns are helpful to stabilize the learning rather than the sparse word-level topics.
\begin{table*}
\setlength{\abovecaptionskip}{0pt}
\setlength{\belowcaptionskip}{0pt}
\caption{\label{tb1}Ten randomly selected topic divergence values at ten successive training epoches with PPMI and NMF, respectively, each being calculated by comparing the context and the predicted response. The variance values are calculated using the ten topic divergence values of PPMI and NMF, respectively.}
\centering
\begin{spacing}{1.29}
\begin{tabular}{c|cccccccccc|c}
 & \multicolumn{10}{c|}{Divergence} & Variance \\
\hline
PPMI & 6.8266e-06 & 0.0916 & 0.0739 & 0.0047 & 0.1112 & 0.0896 & 0.0256 & 0.0008 & 0.0881 & 0.1065 & 0.001897 \\
\hline
NMF & 0.0133 & 0.0106 & 0.0102 & 0.0035 & 0.0093 & 0.0088 & 0.0082 & 0.0010 & 0.0092 & 0.0107 & 0.000012 \\
\hline
\end{tabular}
\end{spacing}
\end{table*}


\subsubsection{Learning local topics}
The local topic distribution is encoded in a dense topic matrix. Topic information is sampled from the dense topic matrix conditioned on the tokens of the contextual utterances. In particular, we match each word in the context with row values of the topic matrix and sum up all words' topic values along the topic dimension in the column. The result value is scaled by the number of tokens to avoid favoring long sentences. Then we get a $d_t$-length vector where $d_t$ is the number of columns in the topic matrix. This vector encodes topics of the context in current dialog. The topic information is dynamic while the conditional factor of the context changes with different dialogs. In the meantime, a topic vector of the predicted response is computed. We use the KL divergence function to measure the difference of the two topic vectors. It is formulated as follows:
\begin{equation}\label{eq4}
  Obj_{local}=KL[tc||tr]=\frac{1}{d_t} \sum_{i=1}^{d_t}KL[tc^i||tr^i]
\end{equation}

By minimizing this objective function, the model is inclined to learning the topic distribution, bringing the final generation and the context closer together in terms of topics. Thus, the generation is not only simply diversified, but also informative and topic-related.

\section{Experimental settings}
\subsection{Datasets}
We conduct experiments on two multi-turn dialog datasets with different styles: Daily Dialogs \cite{li2017dailydialog} and Ubuntu Dialogs \cite{lowe2015ubuntu}. The Daily Dialog corpus contains 13118 high-quality dialogs which are human-written and less noisy. The Ubuntu Dialog corpus has been widely used in multi-turn dialog tasks \cite{serban2016building,serban2017hierarchical,serban2016generative,shen2017conditional}. It consists of almost one million conversations from the Ubuntu chat logs, used to receive technical support for various Ubuntu-related problems. These conversations are arbitrary and lack syntactical regularities. We preprocessed the two datasets, splitting them into three groups of Train, Validation and Test, respectively. Table \ref{tb2} provides descriptive statistics about the two datasets.
\begin{table*}
\setlength{\abovecaptionskip}{0pt}
\setlength{\belowcaptionskip}{0pt}
\caption{\label{tb2}Dataset statistics including number of dialogues in training, validation and test sets, average number of utterances, average number of words per dialogue, and vocabulary size.}
\centering
\begin{spacing}{1.29}
\begin{tabular}{m{2.4cm}<{\centering}m{1.8cm}<{\centering}m{1.8cm}<{\centering}m{1.8cm}<{\centering}m{1.8cm}<{\centering}m{1.8cm}<{\centering}m{1.8cm}<{\centering}}
Corpus & \#Train & \#Validation & \#Test & \#Avg. Utterances & \#Avg. Words & \#Vocab size \\
\hline
Daily Dialogs & 11118 & 1000 & 1000 & 8.9 & 114.7 & 26987 \\
\hline
Ubuntu Dialogs & 448833 & 19584 & 18920 & 7.48 & 102.21 & 268487 \\
\hline
\end{tabular}
\end{spacing}
\end{table*}

\subsection{Baselines}
In the experiments, we evaluate the performance of the proposed model (\textbf{THRED}) against three state-of-the-art neural dialog models, including \textbf{SEQ2SEQ} \cite{vinyals2015neural}, \textbf{HRED} \cite{serban2016building} and \textbf{VHRED} \cite{serban2017hierarchical} which have been discussed in Section Related Work.

\subsection{Metrics of TopicDiv and F Score}
We evaluate the above four models (including the proposed model) from three aspects: producing accurate replies, diversifying the generated responses and generating topic-related responses. The three aspects are demonstrated by the metrics of \emph{Perplexity} \cite{ney1994structuring}, \emph{Distinct} \cite{li2015diversity} and \emph{TopicDiv}, respectively. Taken together, these metrics demonstrate how well the model predicts diversified, informative and topic-coherent responses.

\textbf{\emph{Perplexity}} shows how well a probability model predicts a sample. A lower \emph{Perplexity} indicates the model expects to predict a more accurate reply.

\textbf{\emph{Distinct}} reports the degree of consistency of the generated response to the expectation. A higher \emph{Distinct} value indicates a better model in predicting more diversified responses. It has two indicators: \emph{Distinct1} and \emph{Distinct2}. They calculate the number of distinct unigrams and bigrams of the generated response and scale it by the length of the sequence, respectively. In this paper, unigram \emph{Distinct} is denoted as \emph{Dist1}, and bigram \emph{Distinct} is denoted as \emph{Dist2}.

Besides, we propose a topic-related metric which measures the difference of the context and the generated response in a dialog with respect to the topic information. This metric, called \textbf{\emph{TopicDiv}}, demonstrates topic coherence in the conversation. It is calculated by Eq. \ref{eq4}. The lower \emph{TopicDiv}, the better topic coherence of post-response pairs.

In this paper, we aim to generate both diversified and topic-coherent replies. So, we need a comprehensive metric combining the two factors (i.e., \emph{Distinct} and \emph{TopicDiv}) to evaluate models. Specifically, we introduce the \textbf{\emph{F} score} to do the comprehensive evaluation, which is formulated as follows:
\begin{equation}\label{eq5}
  F_{Disti}^\beta=(1+\beta ^2)\frac{Dist_i\cdot (1-TopicDiv)}{\beta ^2\cdot Dist_i+(1-TopicDiv)}
\end{equation}
Where $\beta$ is a pre-defined real number greater than zero. And the subscript $Disti$ refers to unigram ($i=1$) or bigram ($i=2$) of the metric \emph{Distinct}. When $\beta=1$, \emph{Distinct} and \emph{TopicDiv} contribute equally to this synthetic metric; when $\beta<1$, \emph{Distinct} contributes more yet \emph{TopicDiv} contributes less; and when $\beta>1$, \emph{Distinct} contributes less yet \emph{TopicDiv} contributes more. In this paper, we evaluate models with $\beta=1$, $\beta=0.5$ and $\beta=1.5$, respectively. The higher $F$ score, the better both diversification and topic coherence.

\subsection{Training settings}
The four models including the proposed model (THRED) are all encoder-decoder models. We use the bidirectional LSTM as the encoder part and the unidirectional LSTM as the decoder part. All models have the dimensional size of 500 in the hidden layers. The size of the latent variable $z$ is $d_z=100$. The size of the dense topic features in the (NMF) dense topic matrix is $d_t=40$. For each dataset, we pick top 20000 frequent tokens to make the vocabulary. We train the models with the learning rate of 0.0002. The best validated networks are saved in 400000 training epochs. We also improve the results using Beam Search \cite{sammut2010beam} which samples best-first candidate tokens at each inference step. And we set the Beam number as 5.


\section{Experimental results}
\begin{figure*}[htbp]
\centering
\subfigure[\emph{F} scores on Ubuntu Dialogs with unigram diversification.]{
\includegraphics[scale=0.6]{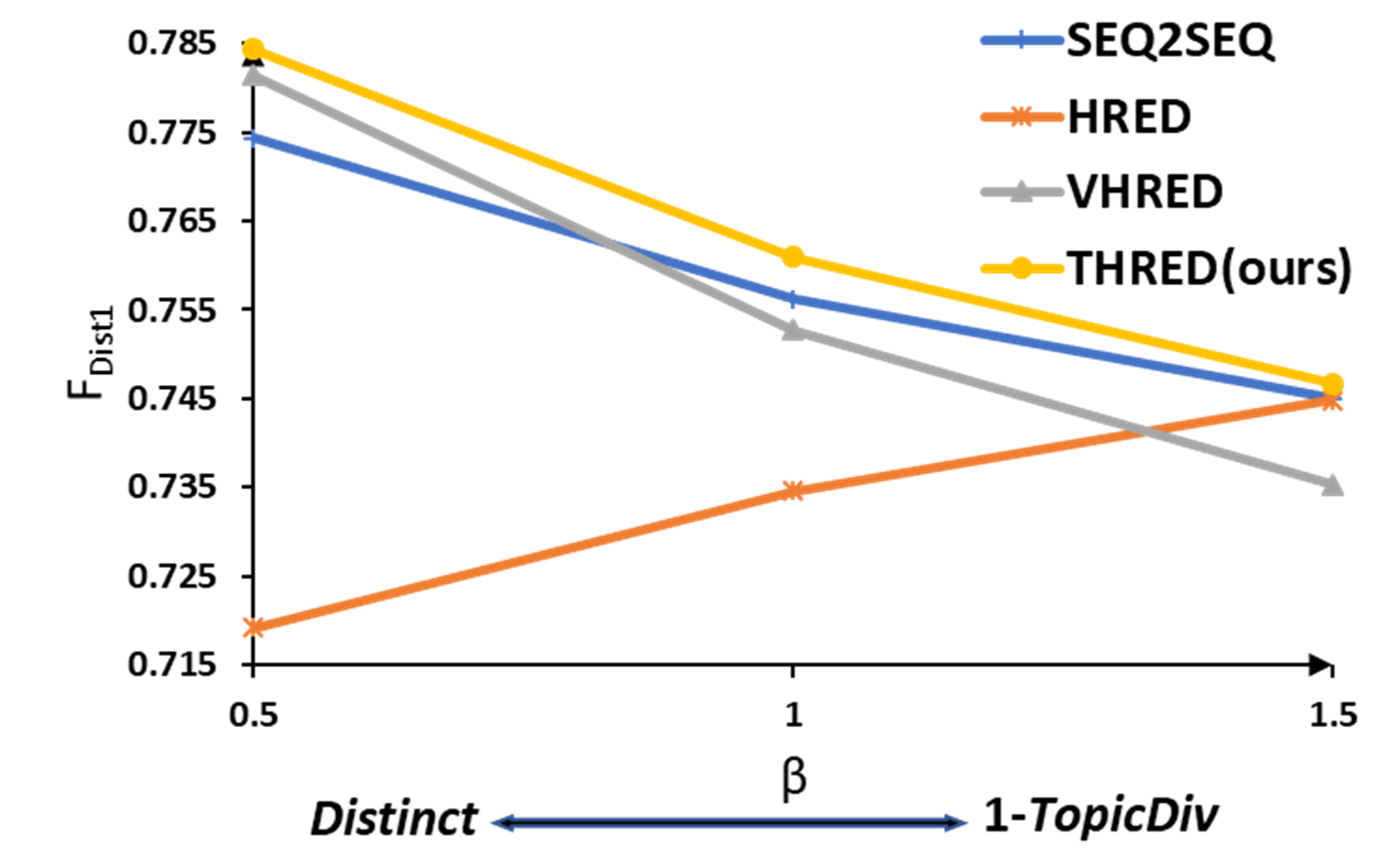}
}
\quad
\subfigure[\emph{F} scores on Daily Dialogs with unigram diversification.]{
\includegraphics[scale=0.6]{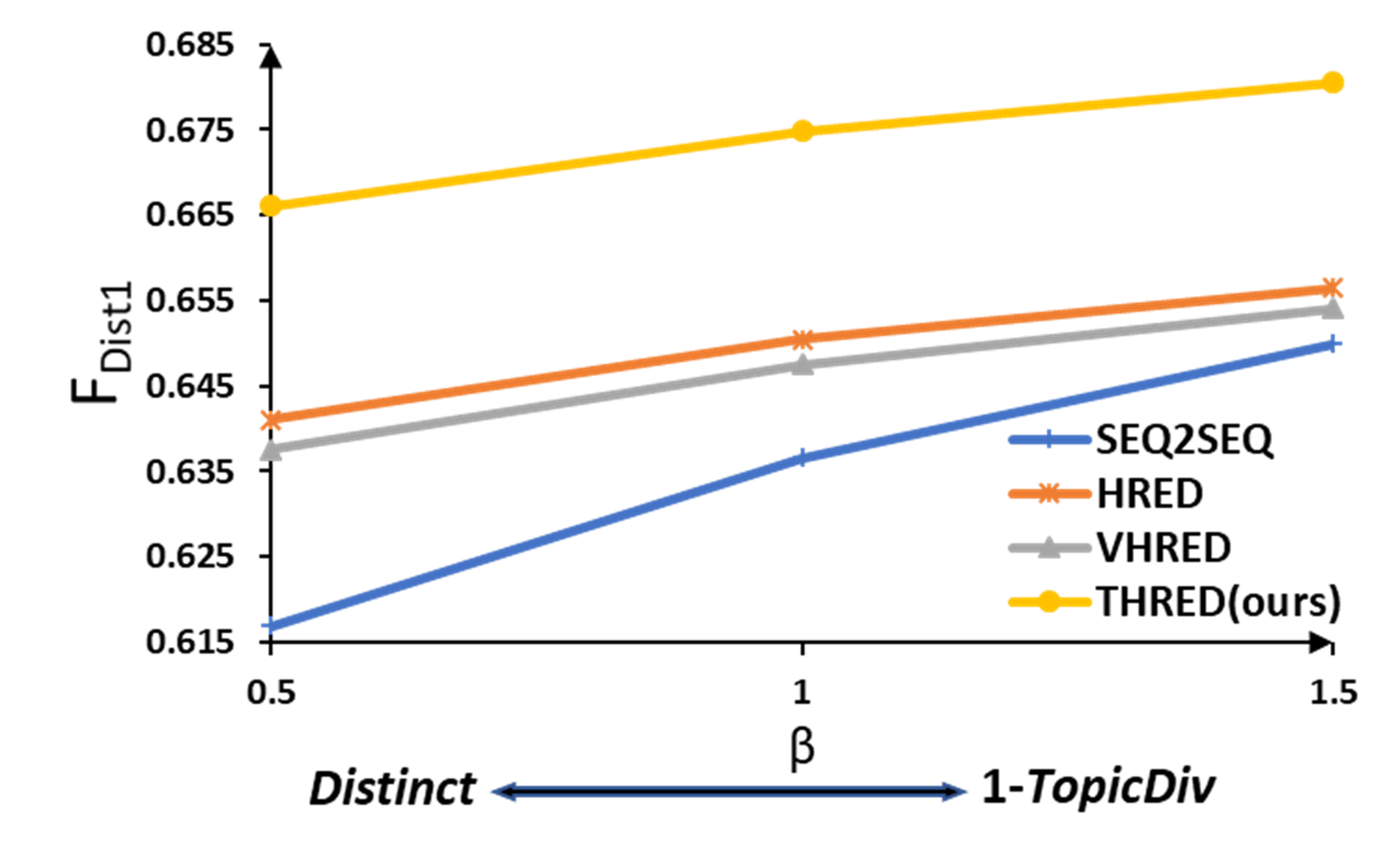}
}
\quad
\subfigure[\emph{F} scores on Ubuntu Dialogs with bigram diversification.]{
\includegraphics[scale=0.6]{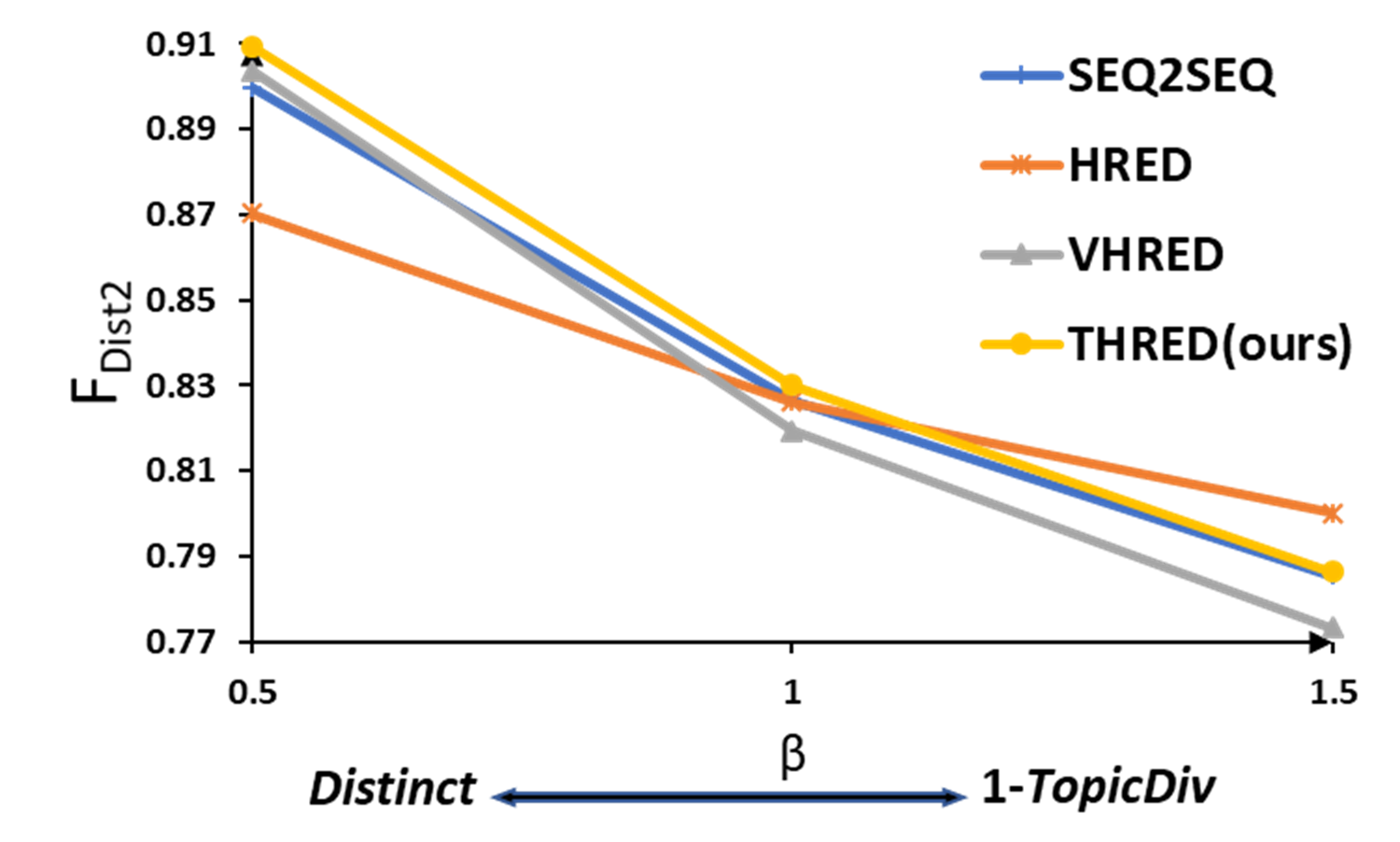}
}
\quad
\subfigure[\emph{F} scores on Daily Dialogs with bigram diversification.]{
\includegraphics[scale=0.6]{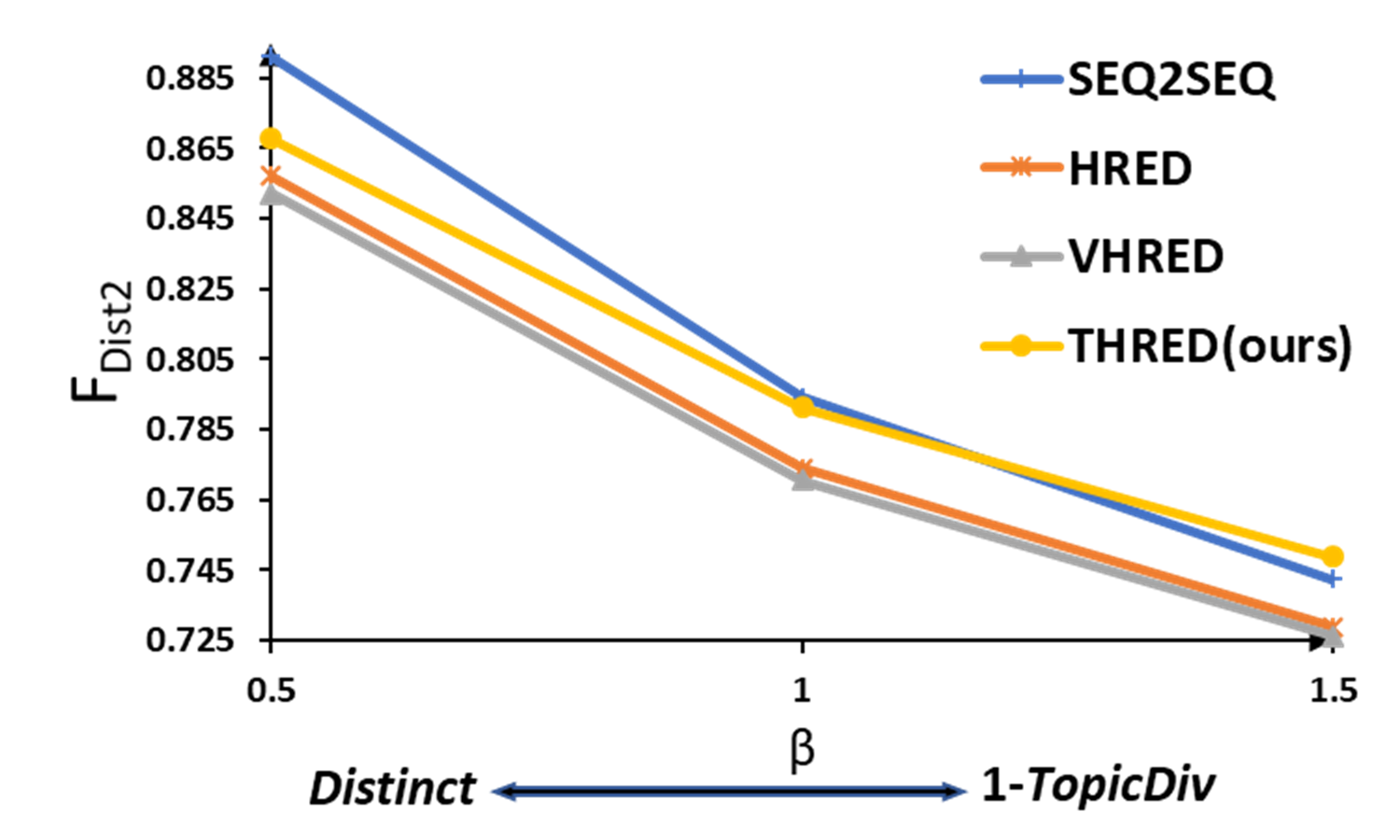}
}
\caption{ \emph{F} scores on respective datasets of Ubuntu Dialogs and Daily Dialogs with unigram or bigram diversification. The vertical coordinates illustrate \emph{F} scores according to respective datasets and uni-(or bi-)gram diversification. The horizontal coordinates denote Diversification-Topic offsets where the lower values represent Diversification-biased \emph{F} scores and the higher values represent Topic-biased \emph{F} scores. }\label{fig2}
\end{figure*}

Evaluation results on datasets of Ubuntu Dialogs and Daily Dialogs are listed in Table \ref{tb3} and Table \ref{tb4}, respectively. We also illustrate the results of the comprehensive metric of \emph{F} scores in Fig. \ref{fig2}, which depicts four sub-figures for the two datasets with unigram diversification (\emph{Dist1}) and bigram diversification (\emph{Dist2}), respectively.

On both datasets of Ubuntu Dialogs and Daily Dialogs, VHRED and THRED perform fairly poor with higher \textbf{\emph{Perplexity} scores}. The reason, we conjecture, is caused by the NLG diversification. When diversifying the generated replies, i.e., replacing tokens and patterns of the expected references with semantically similar ones, the NLG accuracy decreases due to the lack of the tokens of the reference replies. In other words, higher \emph{Perplexity} scores reflect better diversification to some extent.

THRED has much better \textbf{\emph{Dist1} and \emph{Dist2} scores}. Though, on the dataset of Daily Dialogs, SEQ2SEQ achieves the highest \emph{Dist2} score, it performs remarkably worse than the other three models in terms of \emph{Dist1}. And on the dataset of Ubuntu Dialogs, SEQ2SEQ performs much worse than VHRED and THRED in both \emph{Dist1} and \emph{Dist2}. On the other hand, comparing to HRED and VHRED, THRED significantly outperforms HRED on both datasets, performing much better than VHRED on Daily Dialogs, and obtaining fairly equivalent diversification scores to VHRED on Ubuntu Dialogs. In general, THRED has a stable diversification performance, obtaining fairly better diversification scores.

Diversifying NLG leads to the lack of topic coherence of generated replies. As we can see, VHRED performs extremely bad with the highest \textbf{\emph{TopicDiv} scores} as it generates much more diverse replies. However, on the basis of successfully diversifying NLG, THRED performs well in terms of \emph{TopicDiv}, even obtaining the best on the dataset of Daily Dialogs.

Diversifying NLG is not simply seeking substitutes for tokens of the expected reference replies, but replacing them with topic-coherent ones. It is hard to analyze the diversification effect with both \emph{Distinct} (including \emph{Dist1} and \emph{Dist2}) and \emph{TopicDiv} as they are two opposite indicators. In this paper, we advocate \emph{F} score to evaluate models, combining both \emph{Distinct} and \emph{TopicDiv} scores. As shown in the results of Table \ref{tb3} and Table \ref{tb4}, THRED performs rather better with higher \textbf{\emph{F} scores}. In particular, for the unigram diversification, THRED performs better with the highest \emph{F} scores on both datasets and in all situations of Diversification-Topic offsets (w.r.t $\beta=0.5$, $\beta=1$ and $\beta=1.5$). On the other hand, for the bigram diversification, THRED performs better with Diversification-Topic equivalence ($\beta=1$) and Diversification offset ($\beta=0.5$) on the dataset of Ubuntu Dialogs, and performs better than VHRED and HRED in all situations of Diversification-Topic offsets on both datasets. In Fig. \ref{fig2}, it depicts \emph{F} scores according to different Diversification-Topic offsets. As we can see, THRED performs best with unigram diversification and fairly well with bigram diversification.

In overall, THRED performs stably with both fairly better \emph{Distinct} scores and better \emph{TopicDiv} scores. Comparing to the state-of-the-art diversification model of VHRED, THRED improves it with higher \emph{F} scores, especially increasing topic coherence without spoiling the NLG diversification effect.

\begin{table*}
\setlength{\abovecaptionskip}{0pt}
\setlength{\belowcaptionskip}{0pt}
\caption{\label{tb3}Results in terms of accuracy ($Perplexity$), diversification ($Dist1$ and $Dist2$), topic divergence ($TopicDiv$) and \emph{F} scores ($F_{Dist1}$ and $F_{Dist2}$) on Ubuntu Dialogs corpus.}
\centering
\begin{spacing}{1.29}
\begin{tabular}{m{1.4cm}<{\centering}m{1.4cm}<{\centering}m{1.2cm}<{\centering}m{1.2cm}<{\centering}m{1.2cm}<{\centering}m{1.2cm}<{\centering}m{1.2cm}<{\centering}m{1.2cm}<{\centering}m{1.2cm}<{\centering}m{1.2cm}<{\centering}m{1.2cm}<{\centering}}
Model & $Perplexity$ & $Dist1$ & $Dist2$ & $TopicDiv$ & $F_{Dist1}^1$ & $F_{Dist2}^1$ & $F_{Dist1}^{0.5}$ & $F_{Dist2}^{0.5}$ & $F_{Dist1}^{1.5}$ & $F_{Dist2}^{1.5}$ \\
\hline
SEQ2SEQ & 38.1559 & 0.7870 & 0.9564 & 0.2723 & 0.7562 & 0.8265 & 0.7744 & 0.8998 & 0.7450 & 0.7855 \\
\hline
HRED & 39.5231 & 0.7093 & 0.9025 & 0.2382 & 0.7346 & 0.8262 & 0.7192 & 0.8704 & 0.7448 & \textbf{0.8002} \\
\hline
VHRED & 40.6731 & 0.8018 & 0.9702 & 0.2908 & 0.7527 & 0.8194 & 0.7814 & 0.9037 & 0.7353 & 0.7732 \\
\hline
THRED (ours) & 40.7888 & 0.8008 & 0.9712 & 0.2750 & \textbf{0.7610} & \textbf{0.8302} & \textbf{0.7844} & \textbf{0.9094} & \textbf{0.7467} & 0.7863 \\
\hline
\end{tabular}
\end{spacing}
\end{table*}

\begin{table*}
\setlength{\abovecaptionskip}{0pt}
\setlength{\belowcaptionskip}{0pt}
\caption{\label{tb4}Results in terms of accuracy ($Perplexity$), diversification ($Dist1$ and $Dist2$), topic divergence ($TopicDiv$) and \emph{F} scores ($F_{Dist1}$ and $F_{Dist2}$) on Daily Dialogs corpus.}
\centering
\begin{spacing}{1.29}
\begin{tabular}{m{1.4cm}<{\centering}m{1.4cm}<{\centering}m{1.2cm}<{\centering}m{1.2cm}<{\centering}m{1.2cm}<{\centering}m{1.2cm}<{\centering}m{1.2cm}<{\centering}m{1.2cm}<{\centering}m{1.2cm}<{\centering}m{1.2cm}<{\centering}m{1.2cm}<{\centering}}
Model & $Perplexity$ & $Dist1$ & $Dist2$ & $TopicDiv$ & $F_{Dist1}^1$ & $F_{Dist2}^1$ & $F_{Dist1}^{0.5}$ & $F_{Dist2}^{0.5}$ & $F_{Dist1}^{1.5}$ & $F_{Dist2}^{1.5}$ \\
\hline
SEQ2SEQ & 36.8831 & 0.6044 & 0.9699 & 0.3276 & 0.6366 & \textbf{0.7942} & 0.6169 & \textbf{0.8911} & 0.6499 & 0.7425 \\
\hline
HRED & 39.3921 & 0.6349 & 0.9229 & 0.3334 & 0.6504 & 0.7741 & 0.6410 & 0.8570 & 0.6565 & 0.7289 \\
\hline
VHRED & 41.4817 & 0.6310 & 0.9165 & 0.3351 & 0.6475 & 0.7707 & 0.6375 & 0.8520 & 0.6541 & 0.7262 \\
\hline
THRED (ours) & 43.4796 & 0.6604 & 0.9273 & 0.3101 & \textbf{0.6748} & 0.7912 & \textbf{0.6661} & 0.8676 & \textbf{0.6805} & \textbf{0.7489} \\
\hline
\end{tabular}
\end{spacing}
\end{table*}

\section{Discussing diversity}
\emph{Safe reply} has been a long-troubling issue in NLG. It is also stunting the development of RG.

\begin{figure*}
  \centering
  \includegraphics[scale=0.5]{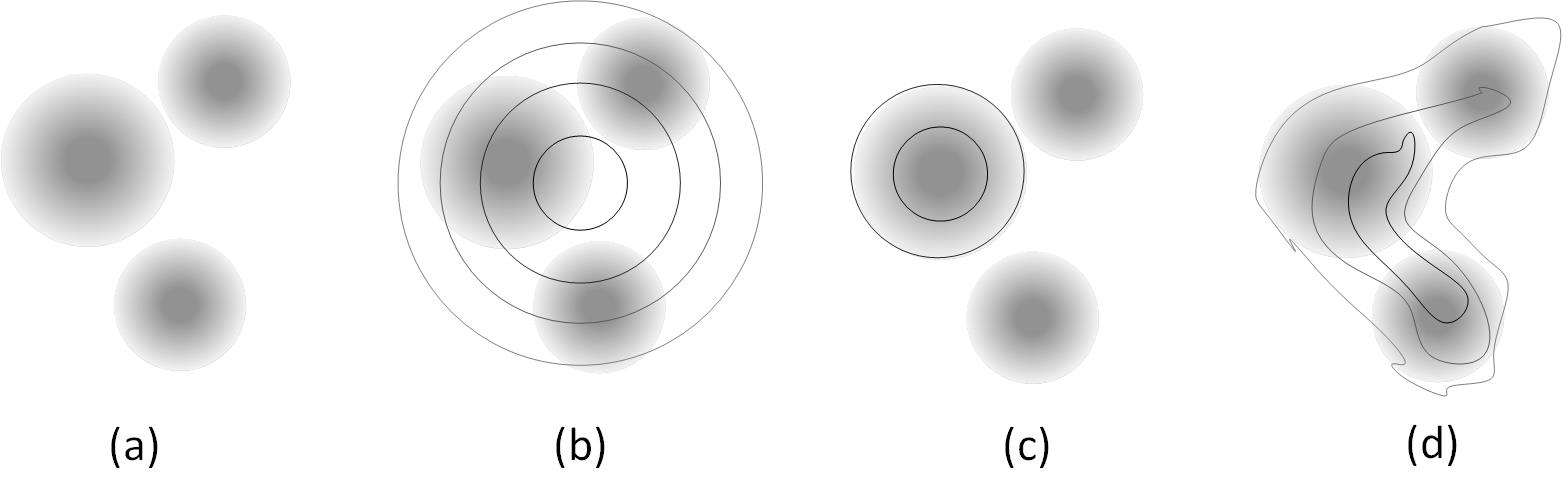}\\
  \caption{Illustration of multi-mode distribution of natural language. (a) A three-mode distribution represents natural language. (b) The coarsely trained system has learned all modes of natural language, yet covering a large white space that denotes patterns outside natural language distribution. (c) The finely trained system has learned a uni-mode distribution of natural language. (d) The well trained system has learned a multi-mode distribution of natural language without deviating too much. }\label{fig3}
\end{figure*}

Natural language presents multi-mode distribution. For the sake of simplicity, we illustrate the multi-mode distribution with three modes and depict it in Figure \ref{fig3} (a). In practice, the system inclines to learn a single-mode distribution \cite{huszar2015not}. The reason, we conjecture, is brought by the gradient-optimizing mechanism of neural networks. Neural networks predict the next token by distributing latent probabilities over all tokens in the vocabulary. And the learning process is performed by driving the probabilities towards the expected tokens. It is formulated as follows:
\begin{equation}\label{eq6}
  CE=\int y\log \hat{y}
\end{equation}

$CE$ is the cross-entropy objective function which has been prevalently used to optimize neural networks \cite{dykeman2016cvae}. $y$ is the expected token and $\hat{y}$ is the predicted token. The optimization of $CE$ aims to push the prediction close to the expectation. The drawback with this process is: The system is trained to produce high-frequent tokens and patterns because they have more chances to show themselves in the optimization process. When the model is finely trained, the frequent patterns force it to cover a single mode (see Fig \ref{fig3} (c)), producing accurate responses. On the contrary, when the model is coarsely trained, although it might cover all modes (see Fig \ref{fig3} (b)), it produces meaningless even ungrammatical sentences due to unpleasant occupation of the learned distribution in the white space \cite{caccia2018language}.

To learn multiple modes of natural language (see Fig \ref{fig3} (d)), we need a dynamical learning strategy. A tradeoff among different learning objectives was created for the dynamic quality \cite{li2015diversity,huszar2015not,fedus2018maskgan}. Especially, the conditional mechanism was adopted to learn the distribution diversity w.r.t each learning objective, extremely improving the flexibility and robustness in modeling dynamic language \cite{serban2017hierarchical,serban2016generative}.

Diversifying RG for multi-turn conversations is not only dispersing the language patterns, but most importantly, is generating informative and topic-related responses. Introducing both the dialog-level contextual distribution and the word-level topic distribution influences the learned safe and commonplace patterns, effectively diversifying the final responses.

\section{Discussing Semantically Invalidity of NLG}

\section{Conclusion}
In this paper, we leverage both dialog-level contextual information and word-level topic coherence to propose the model of THRED, which generates not only diversified but also topic-coherent replies for multi-turn conversations. And we propose an explicit metric (\emph{TopicDiv}) to measure the topic divergence between the post and according replies. In addition, we combine \emph{Distinct} and \emph{TopicDiv} to propose an overall metric which involves both diversification and topic-coherent criteria. We evaluate THRED comparing with three baselines (Seq2Seq, HRED and VHRED) on two real-world corpora, respectively. The results demonstrate that our model performs fairly better for both diversified and topic-coherent response generation.


%

\appendix[Analysis of Generated Replies]
We select ten generated replies according to ten respective contexts and list them in Table \ref{tb5}. We also group these replies (w.r.t. the model of THRED) in two classes: 1) good (topic-coherent and interesting) replies which are listed in items from Item 1 to Item 7, 2) bad (semantically invalid) replies which are listed in items from Item 8 to Item 10. In this Section, we analyze these replies in more details.

In the first three items (Item 1, Item 2 and Item 3), THRED produces diversified alternatives which are not only topic-coherent, but more importantly, proposing specific solutions. For example, in Item 1, ``handbrake'' is an open source application soft for video transcoding, and the context presents issues of how to install and how to make it work. The other four replies (including GrT) propose generic or semantically invalid responses while THRED tries to figure out the issues in a specific way where ``libdvdcss2'' is a lib (supporting) file which could be used to solve certain problems of ``handbrake''\footnote{The ``handbrake'' app is available at https://handbrake.fr/. When ``handbrake'' does not work, throwing out errors such as ``Could not read DVD. This may be because the DVD is encrypted and a DVD descryption library is not installed.'', ``libdvdcss2'' could be an appropriate solution.}.

In Item 4 and Item 5, THRED provides contradictory but valid answers, diversifying the responses without spoiling the semantic consistency to the context. For example, in Item 4, THRED answers by expressing a negative attitude while the other models (including the ground truth) have a grateful expression.

In Item 6 and Item 7, THRED generates semantically equivalent but diversified replies. And in Item 7, both GrT and THRED have proposed specific solutions, giving concrete implementation.

On the other hand, THRED also produces various bad replies. In Item 8, THRED produces an semantically invalid reply as ``usb flash drive'' (of which ``usb'' and ``drive'' appear in the context) is a frequent pattern in the training data \footnote{In the training data, ``usb flash drive'' appears 246 times.}. In Item 9, THRED produces a generic answer responding ``Thank you'' in the context. In Item 10, the context prompts a ``font'' problem; however, THRED mistakes it as a ``screen resolution'' issue due to the keywords of ``monitor'' and ``HD 1920x1080'' in the context.

\begin{table*}
\setlength{\abovecaptionskip}{0pt}
\setlength{\belowcaptionskip}{0pt}
\caption{\label{tb5}The context and its diversified replies. For each context, there are a ground-truth reply (GrT), and four replies produced by four models (THRED, HRED, SEQ2SEQ and VHRED), respectively. THRED generates good (topic-coherent and interesting) replies which are listed in items from Item 1 to Item 7. On the other hand, THRED also generates bad (semantically invalid) replies which are listed in items from Item 8 to Item 10.}
\centering
\begin{spacing}{1.29}
\scriptsize 
\begin{tabular}{m{0.3cm}<{\centering}p{6.6cm}<{\raggedright}p{9.5cm}<{\raggedright}}
\normalsize{Item} & \multicolumn{1}{c}{\normalsize{Context}} & \multicolumn{1}{c}{\normalsize{Replies}} \\
\cline{1-3}
\multirow{7}{*}{\normalsize{1}}
 & \multirow{7}{6.6cm}{\ldots \\
i can find \textbf{handbrake} \dots shows `` Conflicts : handbrake '' \ldots \\
\ldots did you have \textbf{handbrake} installed already \ldots try sudo apt-get install handbrake in a terminal\/ \\
\ldots \\
did you try the terminal . $\land\land\land$ \\
 the actual name is \textbf{handbrake-gtk} and that installs via terminal
} &
\\
&&GrT: cool . :) \\
&&THRED: do you have \textbf{libdvdcss2} installed ? \\
&&HRED: I'm sorry to see you . \\
&&SEQ2SEQ: yes \\
&&VHRED: I'm sorry , sir . \\ \\
\hline
\multirow{5}{*}{\normalsize{2}}
 & \multirow{5}{6.6cm}{\ldots \\
 Are you looking at your computer 's \textbf{CPU temperature} ?\\
 \ldots \\where do i go to \textbf{see its temp} rypervenche ?} &
GrT: That depends on your hardware and kernel . The Ubuntu guys can help you with that .\\
&&THRED: Look at the \textbf{`` sensors '' package} . \\
&&HRED: I don't know , I don't know much about \textbf{lmsensors} , but I don't know much about it . \\
&&SEQ2SEQ: **unknown**  \\
&&VHRED: It should be under System $>$ Administration $>$ \textbf{Power management} . \\
\cline{1-3}
\multirow{5}{*}{\normalsize{3}}
 & \multirow{5}{6.6cm}{anyone ? having trouble \textbf{opening an executable file} , it doesnt do anything ... \\
 no . Be more specific .
} &
GrT: download that game , unzipped , clicked the executable , nothing happens \\
&&THRED: when i try to open a file , it says `` \textbf{no such file or directory} '' \\
&&HRED: i just want to run a program from the terminal , and i want to run it from the terminal , but i don't know how to do that \\
&&SEQ2SEQ: what do you mean ? \\
&&VHRED: trying to open a file , it says it \textbf{can't find the file} \\
\cline{1-3}
\multirow{5}{*}{\normalsize{4}}
 & \multirow{5}{6.6cm}{Does anyone have any experience setting up Ubuntu with MAAS ? I'm having some weird time issues when trying to add nodes . \\
 The folks in \#ubuntu-server might be of some help
} &
GrT: Thanks much . \\
&&THRED: I've tried that , but I can't seem to find a solution . \\
&&HRED: Thanks . \\
&&SEQ2SEQ: thanks .  \\
&&VHRED: thanks !  \\
\cline{1-3}
\multirow{5}{*}{\normalsize{5}}
 & \multirow{5}{6.6cm}{how can I remove a launchpad keyring , \ldots \\
 apt-key del 1378B444
} &
GrT: thanks :) \\
&&THRED: I tried that , but it didn't work \\
&&HRED: thanks \\
&&SEQ2SEQ: thanks \\
&&VHRED: I tried that , but it still doesn't work \\
\cline{1-3}
\multirow{5}{*}{\normalsize{6}}
 & \multirow{5}{6.6cm}{Hi , does anyone know of any \textbf{file shredding software} for Ubuntu ? \\
 there is shred , but it doesn't really work on journaling filesystems .
} &
GrT: i'm on ext4 would that work ? \\
&&THRED: do you know of any software that can \textbf{recover deleted files} ? \\
&&HRED: ok , thanks \\
&&SEQ2SEQ: **unknown** , **unknown** , **unknown** , **unknown** , **unknown** \\
&&VHRED: **unknown** , that 's what I was looking for , thanks \\
\hline
\multirow{6}{*}{\normalsize{7}}
 & \multirow{6}{6.6cm}{hi ! ... i have \textbf{changed the default zoom level} and now i cant go back to the default or ` 66\% ' ... can any one help me here plz ? \\
\ldots
} &
GrT: hm . Hold down Control key while scrolling the mousewheel ? That 's a guess , not an answer . \\
&&THRED: ok , open a terminal and type : \textbf{gconf-editor} \\
&&HRED: are you using compiz ? \\
&&SEQ2SEQ: what do you mean ? \\
&&VHRED: I'm not sure what you mean by that . I don't know how to fix it though . \\

\cline{1-3}
\multirow{6}{*}{\normalsize{8}}
 & \multirow{6}{6.6cm}{HEy all  I don't have any dvds at the moment  Can I load an external hdd with ubuntu then use that to install it ?  \ldots is that possible to do with this \textbf{hardrive} ? a hard drive is \textbf{hard drive} isn't it ? \ldots \\
 no \textbf{usb} thumb ? \\
 nope
} &
GrT: http://ubuntuforums.org/showthread.php?t=1549847 no problem you can boot the iso from grub \\
&&THRED: it 's a \textbf{usb flash drive} \\
&&HRED: do you have a \textbf{usb flash drive} ? \\
&&SEQ2SEQ: it 's a usb drive , it 's a \textbf{usb drive} \\
&&VHRED: then you should be able to boot from \textbf{usb} \\
\cline{1-3}
\multirow{7}{*}{\normalsize{9}}
 & \multirow{7}{6.6cm}{\ldots \\
 \ldots more than 2GB you might see advantages using the 64-bit installation . \ldots \\
 \ldots \\
 \ldots we can install the 32-bit libraries to support 32-bit applications on 64-bit without any silly kludges we had before \\
 excellent . That 's good to know - reassuring . Thank you . :)
 } &
\\
&&GrT: check the specifications of the manufacturer \\
&&THRED: You're welcome . \\
&&HRED: no problem , glad to help \\
&&SEQ2SEQ: it 's a bit of a pain in the ass , but it 's a bit of a **unknown** . \\
&&VHRED: you're welcome \\
\\
\cline{1-3}
\multirow{10}{*}{\normalsize{10}}
 & \multirow{10}{6.6cm}{\ldots
  When I log in with Awesome as my WM , the fonts in the Terminal , \ldots , have an EXTREMELY small font \ldots and the fonts when using Gnome are perfect . \\
 \ldots use the key board \\
 \ldots \\
 yes there is Alt+f2 or ctrl+alt+f1 does the keyboard work ? \ldots \\
 I don't know if this has anything to do with it but my \textbf{monitor} is full \textbf{HD 1920x1080} , 47 '' \ldots \\
 \ldots
 } &
\\\\
&&GrT: plugged to the pc via HDMI \\
&&THRED: I am trying to figure out how to change the \textbf{resolution of my monitor} \\
&&HRED: I think it 's a font problem , but I don't know how to do that . \\
&&SEQ2SEQ: what do you mean ? \\
&&VHRED: I have a laptop , and it 's a laptop . \\
\\\\\\
\hline
\end{tabular}
\end{spacing}
\end{table*}

\section*{Acknowledgment}

The authors would like to thank...

\section*{Conflict of interest}

 The authors declare that they have no conflict of interest.

\ifCLASSOPTIONcaptionsoff
  \newpage
\fi



\bibliographystyle{IEEEtran}
\bibliography{IEEEabrv,mybibfile}
%
%
%

%

\begin{IEEEbiography}{Michael Shell}
Biography text here.
\end{IEEEbiography}

\begin{IEEEbiographynophoto}{John Doe}
Biography text here.
\end{IEEEbiographynophoto}


\begin{IEEEbiographynophoto}{Jane Doe}
Biography text here.
\end{IEEEbiographynophoto}




\end{document}